# Multi-document Summarization using Semantic Role Labeling and Semantic Graph for Indonesian News Article

Yuly Haruka Berliana Gunawan
School of Electrical Engineering and Informatics
Bandung Institute of Technology
Bandung, Indonesia
yulyharuka@gmail.com

Masayu Leylia Khodra
School of Electrical Engineering and Informatics
Bandung Institute of Technology
Bandung, Indonesia
masayu@stei.itb.ac.id

*Abstract*—In this paper, we proposed a multi-document summarization system using semantic role labeling (SRL) and semantic graph for Indonesian news articles. In order to improve existing summarizer, our system modified summarizer that employed subject, predicate, object, and adverbial (SVOA) extraction for predicate argument structure (PAS) extraction. SVOA extraction is replaced with SRL model for Indonesian. We also replace the genetic algorithm to identify important PAS with the decision tree classifier since the summarizer without genetic algorithm gave better performance. The decision tree model is employed to identify important PAS. The decision tree model with 10 features achieved better performance than decision tree with 4 sentence features. Experiments and evaluations are conducted to generate 100 words summary and 200 words summary. The evaluation shows the proposed model get 0.313 average ROUGE-2 recall in 100 words summary and 0.394 average ROUGE-2 recall in 200 words summary.

*Keywords—multi-document summarization, semantic role labeling, decision tree classification model.*

## I. Introduction

Nowadays, a lot of news sources can be found on the internet. There is a lot of news articles on the internet just for one topic. However, each news article has its viewpoint and has a great amount of possibility that it does not provide complete information about the topic. Other than that, if people read more than one news article about one topic, there is a lot of redundant information provided in each news article. Therefore, an automatic news summarization technique may overcome these problems.

There are two types of automatic summarization techniques according to its result. First, extractive summarization selects important text units and merge them into a summary. Second, abstractive summarization executed by paraphrase the extracted important information to generate a summary. The extractive technique is much more widely used because it is simpler than the abstractive summarization technique.

Summarization has two approaches to be employed i.e. syntactic approach and semantic approach [2]. Syntactic approach uses syntactic parser to analyze the sentence from its syntax, and semantic approach uses semantic representation of the sentence to analyze it.

Khan et al. [2] developed multi-document summarization using semantic approach. Summary from multi-document news articles is generated by employing predicate argument structure (PAS) extraction, genetic semantic graph, and natural language generation (NLG) [2]. Khan et al. [2] outperformed the best summarization model for the DUC 2002 dataset based on an average of ROUGE-1 and ROUGE-2 score.

Devianti and Khodra [1] adapted Khan et al. [2] architecture to develop a summarizer for Indonesian news articles. This summarizer [1] achieved better performance than Reztaputra and Khodra [3] summarizer. Other than that, this model [2] outperformed Garmastewira and Khodra [4] extractive summarizer for a 200-words summary. Devianti and Khodra [1] employed subject, verb, object, and adverbial (SVOA) extraction component as PAS extraction components because there are no SRL tools for Indonesian sentence that has been proven good. This SVOA extraction can only recognize two adverbials, i.e temporal and location adverbial. This limitation makes the calculation of semantic similarities between each PASs not taking into account other adverbials besides temporal and location. This can conduct redundant information in the generated summary because the sentence in the Indonesian language news has more varied structure so that there are other adverbials which is also important, such as the cause, the effects, and so forth.

Meanwhile, Devina and Khodra [5] developed SRL for Indonesian to generate Indonesian news template. However, this SRL model [5] has limitations. The SRL model is only able to recognize 60 predicates. But from the experiment that has been done by Devina and Khodra [5], this model can predict the predicates outside the 60 predicates. Our proposed model will replace the SVOA extraction component with SRL component to improve summarizer performance.

From Devianti and Khodra [1] experiment, important PAS identifier model using four sentence features with default weight gave better performance than genetic algorithm model to determine PAS features weight. Khan et al. [2] also gets the best results by using the four sentence features but with genetic algorithm. The genetic algorithm is used to search optimal weight for every PAS features so it can improve the performance of the summarizer. Since model without genetic algorithm gave better performance, we employed another algorithm to improve the summarizer performance. Different algorithms have been investigated for sentence features extraction [2, 4, 6]. Hannah et al. [6] use fuzzy inference system to extract sentence features to identify important sentences in the article. We employ a classification model i.e. decision tree, to determine important and not important PAS.

We propose a multi-document summarization model with semantic role labeling and semantic graph using Devianti and Khodra [1] architecture. Modifications were made to Devianti and Khodra [1] model by replacing the SVOA extraction with SRL and genetic algorithm with a classification model to extract sentence features.

Section 2 discusses related works, i.e. abstractive summarization for Indonesian news articles [1], semantic role labeling for Indonesian [5], and sentence classification technique [6, 7]. The proposed methods will be discussed in section 3. The experiments and evaluations will be described in section 4. Finally, section 5 will discuss the conclusions of the result.

## II. RELATED WORKS

### A. Abstractive Summarization using Genetic Semantic Graph for Indonesian News Articles

Devianti and Khodra [1] developed abstractive summarizer architecture using genetic semantic graph for Indonesian news articles. Fig.1 depicts the architecture in pipelines. First of all, the sentence is preprocessed to clean the sentence. Then applied SVOA extraction to each sentence to extract it into predicate argument structure (PAS). Create a semantic similarity matrix using similarity score each pair of PAS that calculated using Word2Vec Skipgram and fastText Kyubyong (OOV). Create semantic graph using semantic similarity matrix with PASs as vertices and similarity score as the weight of the edge. After that, calculate sentence features and calculate the importance score of PASs using each sentence features weight from genetic algorithm from training. Importance score of PASs calculated using modified weighted graph based ranking algorithm (MWGRA). Using maximal marginal relevance (MMR), PASs will be selected based on its importance score. And then, natural language generation (NLG) will be used to create new sentences from the PASs that has been chosen before. Heuristic rules are applied in NLG.

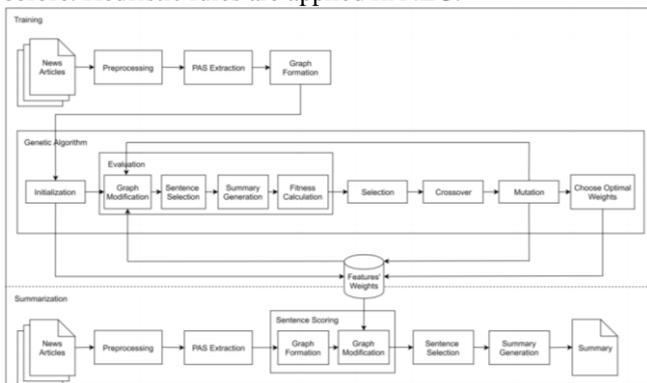

Fig. 1. Abstractive News Summarization for Indonesian News Articles [1]

### B. Semantic Role Labeling for Indonesian

Devina and Khodra [5] developed SRL in Indonesian to generate Indonesian news template. This SRL model consists of two modules, predicate identification module and argument identification module. Fig.2 depicts the architecture. Both of the modules were built using two BiLSTM layers with softmax activation function on the last layer. Predicate identification module is used to identify the predicate in the sentence. The output from the predicate identification module becomes the input in the argument identification module. The argument identification module is used to identify the SRL argument for each word in a sentence according to its relation to the predicate that identified in predicate identification module.

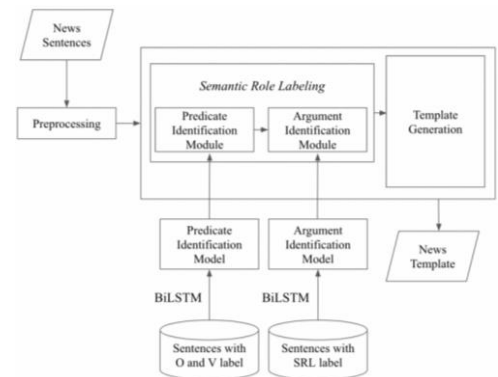

Fig. 2. Semantic Role Labeling for Generating Indonesian News Template [5]

### C. Sentence Classification

Hannah et al. [6] developed an extractive summarizer using fuzzy inference system to extract sentence features. This model [6] has seven sentence features and extracts it into importance score using fuzzy inference system. The importance score classifies the sentence into 3 classes, i.e important, average, and unimportant. This summarization model [6] uses fuzzy rules. Apparently, Hannah et al [7], did not mention the fuzzy rules in their paper. On the other hand, Kupiec et al. [7] developed a sentence classification model that classifies sentences into probability values whether the sentence is part of a summary based on the value of sentence features. Therefore, the sentence classification model can be used as an alternative to genetic algorithms in Devianti and Khodra [1] model.

## III. PROPOSED METHOD

We proposed multi-document summarization by modifying Devianti and Khodra [1] architecture. The modifications are made by replacing SVOA extraction into SRL and replacing genetic algorithm with a classification model. In this model, the decision tree classification model used to replace the genetic algorithm [1]. The decision tree classification model classifies sentences in the document into important and unimportant sentences based on sentence features by Khan et al. [2]. The decision tree model was chosen because it is ideal for discrete selection such as important and unimportant sentences. In addition, this algorithm was chosen because it is ideal for classifying sentences based on the most dominant or most influential sentence features. The semantic graph also used in this model to improve summarizer performance. The semantic graph applied to reduce redundancy in summary. In multi-document summarization, a lot of sentences that are similar or the same in several documents. With the semantic graph, we can reduce redundancy by not connecting the similar sentences in the graph, where if the similarity value between two PASs greater than 0.5, the PASs will not be connecting in the graph. Fig. 3 depicts the architecture of the proposed system. This abstractive summarizer has six main components. In this paper, we discuss about the changes in each component.

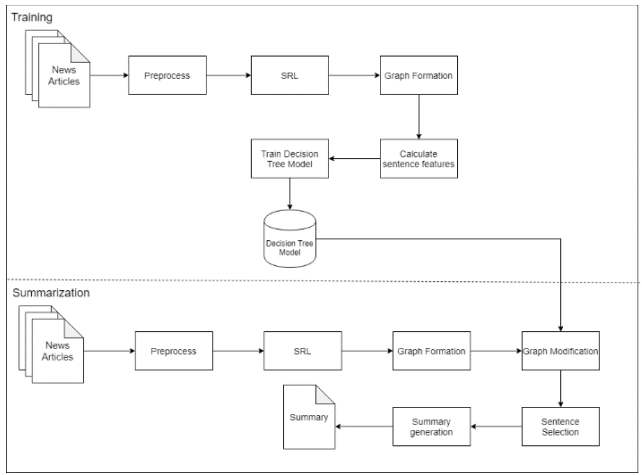

Fig. 3 Proposed abstractive summarization system using semantic role labeling for Indonesian news articles

*A. PAS Extraction using SRL*

In this component, we replace the SVOA extraction [1] using the SRL model [5]. The SRL model can predict 11 arguments including verbs from a sentence [5]. TABLE I shows the 11 arguments.

TABLE I. SRL LABEL

| Argument | Name |
|---|---|
| A0 | *Agent* |
| V | *Verb* |
| A1 | *Patient* |
| AM-LOC | *Annotation Modifier Locatives* |
| AM-TMP | *Annotation Modifier Temporal* |
| AM-GOL | *Annotation Modifier Goal* |
| AM-CAU | *Annotation Modifier Cause Clauses* |
| AM-EXT | *Annotation Modifier Extent* |
| AM-ADV | *Annotation Modifier Adverbials* |
| AM-MOD | *Annotation Modifier Modals* |
| AM-NEG | *Annotation Modifier Negation* |

The SRL model has some limitations. These are the limitations:

1. The SRL model can only recognize 60 predicates that have been identified in the SRL training data. In the summary corpus, there are 76 unique titles and only 37 titles have been identified.

2. The SRL model's performance is still bad when applied to complex sentences.

To overcome the limitations, 2 solution approaches were adopted. The first approach is to add training data for the re-model of the SRL model. The goal of this approach is to make the SRL model recognize more predicates and recognize the correct predicate in complex sentences.

The second approach is to use dependency parser for improve performance in predicate identification since Devianti and Khodra [1] have good performance in predicate identification using dependency parser and rules. We use StanfordNLP and anytree to generate the dependency parser tree and use Devianti and Khodra [1] rules to extract the predicate. Fig. 4 depicts the illustration of the usage of dependency parser in the SRL model.

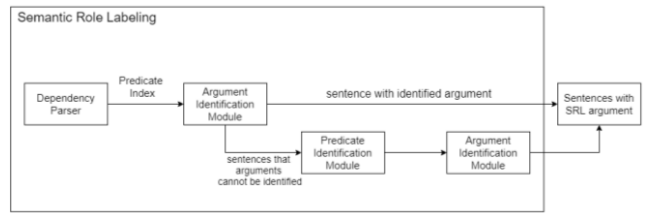

Fig. 4 Proposed SRL Model using Dependency Parser

*B. Sentence Scoring*

In this component, we create a graph using the semantic similarity matrix in graph formation component and modify the weight of the vertices and edges of the graph in graph modification component.

*1) Graph Formation*

First, we calculate the semantic similarity of each pair of PASs using pre-trained word embedding. Based on Devianti and Khodra [1] experiment, Word2Vec SkipGram and fastText Kyubyong generate the best result for Indonesian news articles summarization. So in this model, we will use Word2Vec SkipGram and fastText Kyubyong to calculate the semantic similarity of each pair of PASs. Semantic similarity between two PASs will be measure by calculating the similarity score of each argument. The argument that will be used in this model is 11 SRL argument in TABLE I. The similarity between two words will be calculated using cosine similarity based on the word embedding. And then, the similarity between two PASs will be calculated by taking an average of similarity scores for every argument. Fig. 5 shows the example of using SRL argument in the semantic similarity between *"sebelumnya korban meminta kertas"* (previously the victim asked for a paper) and *"Jasad korban berada di rumah sakit kini"* (the victim's body is in the hospital now).

| Argumen SRL | | A0 | | V | | AM-LOC | | AM-TMP |
|---|---|---|---|---|---|---|---|---|
| | | *Jasad* (body) | *korban* (the victim's) | *berada* (is) | *di* (in) | *Rumah Sakit* (the hospital) | | *kini* (now) |
| A0 | *korban* (the victim) | 0.504 | 1.0 | - | - | - | | - |
| V | *meminta* (asked for) | - | - | -0.01 | - | - | | - |
| A1 | *kertas* (a paper) | - | - | - | - | - | | - |
| AM-TMP | *sebelumnya* (previously) | - | - | - | - | - | | 0.156 |

Fig. 5 Example of Semantic Similarity Calculation

Afterward, a semantic graph is constructed where PASs as vertices of the graph and similarity score of each pair of PASs as the weight of the edge of the graph. Based on [2], PAS will be connected to another PAS if the semantic similarity score between the PASs is in the range of $0 < \alpha \leq 0.5$. This component uses NetworkX to form the semantic graph. Fig. 6 shows the example of the semantic graph of 4 sentences, where the vertex is the PAS and the edge is semantic similarity between PAS

| Semantic Similarity Between PASs | Semantic Graph |
|---|---|
| sim(P1,P2) = 0.265 | |
| sim(P1,P3) = 0.189 | |
| sim(P1,P4) = 0.224 | |
| sim(P2,P3) = 0.391 | |
| sim(P2,P4) = 0.486 | |
| sim(P3,P4) = 0.396 | |

Fig. 6 Example of Semantic Graph

## 2) Graph Modification

First of all, we calculate PAS to document relationships using ten sentence features, i.e PAS to PAS similarity, PAS position, semantic term weight, and frequent semantic term, PAS length, proper nouns, numerics, nouns and verbs, temporals, and locations [1]. Calculations of the PAS to document relationship can be calculated as follows [1]:

$$w(P_i, D_{set}(P_i)) = \sum_{k=1}^{n} w_k * P_{i\_}f_k \quad (1)$$

where $w(P_i, D_{set}(P_i))$ is PAS $P_i$ to document $D_{set}(P_i)$ relationships score, $P_{i\_}f_k$ is the calculated score for feature k obtained from PAS $P_i$, $w_k$ is the weight of the feature k, and n is the number of total features.

Devianti and Khodra [1] use genetic algorithm to define the weight for each of the sentence features. In this system, we will use classification model, especially decision tree learning, to extract the PAS to document relationships. We use sentence features as the features in the model and the output will be '1' or '0', where '1' represents the important sentence in the news article and '0' represents the unimportant sentence in the news article. And $w_k$ for all of the feature is 1.

If the PAS has label '1', new weight for the edges in the semantic graph will be calculated as follows [2]:

$$f(P_i, P_j | D_{set}(P_i), D_{set}(P_j)) = sim_{sem}(P_i, P_j) * [\mu * w(P_i, D_{set}(P_i)) + (1 - \mu) * w(P_j, D_{set}(P_j))] \quad (2)$$

where $w_k$ for all of the feature is 1, and $\mu \in [0,1]$ is the combination weight that controls the contribution of the value of the connectedness of one PAS with another PAS. We use 0.5 which is the optimal value for the $\mu$ [1].

If the PAS has label '0', we use the same equation in [2], but we change the value of $w(P_i, D_{set}(P_i))$ to 0. It means that the PAS has no relationship to the document. We can see the equation as follows [3]:

$$f(P_i, P_j | D_{set}(P_i), D_{set}(P_j)) = sim_{sem}(P_i, P_j) * [\mu * 0 + (1 - \mu) * w(P_j, D_{set}(P_j))] \quad (3)$$

After that, we applied modified weighted graph-based ranking algorithm (MWGRA). We use MWGRA to calculate the ranking of the PAS and the importance score of the PASs by using a new weight of the edges as calculated in equation [2] and equation [3] and added it with the semantic similarity score. Calculation of MWGRA can be seen in [1].

## IV. EXPERIMENTS AND EVALUATION

### A. Experiments

The purpose of our experiments are:

1. To test the quality of the new SRL model with added training data.
2. To determine the best decision tree learning model for summarizing 100 words and 200 words.

For the SRL model experiment we use Devina and Khodra [5] training and added training data as shown in TABLE II.

TABLE II. STATISTIC OF THE TRAINING DATA FOR SRL MODEL

| Label and Parameters | Total Data of Pretrained Training Data SRL | Total Data on SRL Modeling | Total Additional Data |
|---|---|---|---|
| Number of Sentences | 1141 | 2073 | 932 |
| Number of Unique Predicates | 60 | 324 | 264 |
| Number of Words | 11952 | 24787 | 12835 |

We also use Devina and Khodra [5] validation data for the experiment. For the decision tree classification model, we use Devianti and Khodra [1] training and validation data.

Experiment for SRL model results F1 score 1.0 in predicate identification module and F1 score 0.6843 in argument identification module.

The experiment for decision tree classification model uses 40 topics in training corpus for training and validation data. We developed two decision tree models, i.e decision tree model with four sentence features and decision tree model with 10 sentence features. Both of decision tree classifier result in 1.0 for precision and also for recall to determine the label for the sentence. We use both of the models for the experiment to determine the best decision tree model. The decision tree model with 10 sentence features has better performance than the decision tree model with four sentence features both at a 100 word summary and a 200 word summary. Therefore, decision tree with 10 sentence features will be used in the evaluation.

### B. Evaluation

We conduct testing by using corpus from Devianti and Khodra [1] that consists of 5 news topics and each topic has around 10-20 news articles. For the SRL model testing, we use labeled sentences from *"Bunuh Diri"* (Suicide) news articles. The number of sentences in *"Bunuh Diri"* (Suicide) corpus is 114 sentences. TABLE III. shows the result of the SRL model testing. We use the F1 score to compare both models, i.e SRL and SRL with dependency parser. Based on the evaluation, SRL has a better performance than SRL with dependency parser. However, SRL with dependency parser recognizes more predicate in testing data. Therefore, both of the SRL models will be used to determine the best model.

TABLE III. RESULT OF SRL MODEL TESTING

| Argument | SRL F-Measure | *SRL with Dep Parser* F-Measure |
|---|---|---|
| A0 | 0.728 | 0.725 |
| A1 | 0.664 | 0.652 |
| V | 0.649 | 0.648 |
| AM-LOC | 0.810 | 0.803 |
| AM-TMP | 0.631 | 0.646 |
| AM-GOL | 0 | 0 |
| AM-CAU | 0 | 0 |
| AM-EXT | 0 | 0 |
| AM-ADV | 0 | 0 |
| AM-MOD | 0 | 0 |
| AM-NEG | 1 | 1 |

Based on the result of the experiment, ten sentence features decision tree has a better performance than the four sentence features decision tree. In this section, we test if the ten sentence features decision tree model can result in better performance in the summarization task. We test the decision tree model with the new model of SRL without the dependency parser. TABLE IV. show the result of the decision tree model testing.

TABLE IV. RESULT OF DECISION TREE MODEL

| Summary | Average Recall | |
|---|---|---|
| | SRL with *Decision Tree 10* | SRL |
| 100 words summary | 0.25106 | 0.21596 |
| 200 words summary | 0.27946 | 0.18949 |

From the comparison of the testing result, we can see that model with ten sentence features decision tree classification model result in a better performance than no decision tree model. Therefore we will use the ten sentence feature decision tree model to determine the best model.

In this section, we will compare our model with Devianti and Khodra model [1]. We use Devianti and Khodra [1] best model that is a summarization model with 4 sentence feature using default weight in each feature. We also use Devianti and Khodra summarization model using the best model of genetic algorithm [1] to be compared. And for our model, we use SRL with dependency parser for PAS extraction and decision tree classification model for classifying the important sentences. We also compare our model with SRL and decision tree classification model with Devianti and Khodra [1] model.

TABLE V. and TABLE VI. shows the performance comparison of different models.

TABLE V. PERFORMANCE COMPARISON OF 100 WORDS SUMMARY

| Model | Min *Recall* | Max *Recall* | Avg ROUGE-2 *Recall* |
|---|---|---|---|
| dep-parser-srl-dtl | 0.090 | 0.755 | 0.313 |
| srl-dtl | 0.045 | 0.755 | 0.251 |
| Devianti-Khodra-default-4 | 0.034 | 0.670 | 0.320 |
| Devianti-Khodra-best-model | 0.102 | 0.5 | 0.282 |

TABLE VI. PERFORMANCE COMPARISON OF 200 WORDS SUMMARY

| Model | Min *Recall* | Max *Recall* | Avg ROUGE-2 *Recall* |
|---|---|---|---|
| dep-parser-srl-dtl | 0.1 | 0.687 | 0.394 |
| srl-dtl | 0.033 | 0.687 | 0.189 |
| Devianti-Khodra-default-4 | 0.156 | 0.604 | 0.394 |
| Devianti-Khodra-best-model | 0.111 | 0.559 | 0.357 |

From the comparison of the testing result, we can see that dep-parser-srl-dtl has the best performance compared to srl-dtl and Devianti-Khodra-best-model according to average ROUGE-2 recall. However, dep-parser-srl-dtl has worse performance than Devianti-Khodra-default-4 in 100 words summary according to average recall ROUGE-2.

In this section, we will compare our best model summary with Devianti and Khodra best algorithm genetic models summary. TABEL VII. shows the result of the comparison.

TABLE VII. SUMMARIES GENERATED FROM OUR BEST MODEL AND BEST MODEL OF DEVIANTI & KHODRA [1]

| **Our best model (Dep-parser-srl-dtl)** |
|---|
| Indosat Ooredoo bersama Fujitsu Indonesia menandatangani Nota Kesepahaman dalam rangka memantapkan kemitraan dengan para pelanggan bisnis yang telah dibangun dan bekerja sama , dalam menghadirkan solusi Smart Mobility dan Internet of Things ( IoT ) . |
| Sebagai langkah awal , kerjasama ini akan berfokus pada sektor otomotif dan transportasi dan akan diperluas ke berbagai sektor industri termasuk sektor publik untuk memenuhi kebutuhan pelanggan korporasi di Indonesia tetapi juga perusahaan-perusahaan Jepang yang beroperasi di Indonesia . … menghadirkan produk-produk dan layanan baru …. |
| *Indosat Ooredoo and Fujitsu Indonesia signed a Memorandum of Understanding to strengthen partnerships with business customers that have been built and work together, in presenting Smart Mobility and Internet of Things (IoT) solutions. As a first step, this collaboration will focus on the automotive and transportation sectors and will be extended to various industrial sectors including the public sector to meet the needs of corporate customers in Indonesia but also Japanese companies operating in Indonesia.... presenting new products and services ...* |
| **Devianti-Khodra-best-model** |
| …. Indosat Ooredoo dan Fujitsu Indonesia sekaligus bekerjasama dalam menghadirkan solusi Smart Mobility dan Internet of Things ( IoT ) . Indosat Ooredoo bekerja sama dengan Fujitsu menghadirkan solusi Smart Mobility dan Internet of Things ( IoT ) . Kerjasama ini akan berfokus pada sektor otomotif dan transportasi lalu merambat ke berbagai sektor industri termasuk sektor publik dan akan diperluas ke berbagai sektor industri termasuk sektor publik untuk memenuhi kebutuhan pelanggan korporasi di Indonesia tetapi juga perusahaan-perusahaan Jepang yang beroperasi di Indonesia . |
| *Indosat Ooredoo and Fujitsu Indonesia at the same time work together in presenting Smart Mobility and Internet of Things (IoT) solutions. Indosat Ooredoo collaborates with Fujitsu to present Smart Mobility and Internet of Things (IoT) solutions. This collaboration will focus on the automotive and transportation sectors and then spread to various industrial sectors including the public sector and will be extended to various industrial sectors including the public sector to meet the needs of corporate customers in Indonesia but also Japanese companies operating in Indonesia.* |

From the comparison we made, our model results in a better summary by having more phrases or sentences that intersect with a reference summary. Moreover, the order of information in the dep-parser-srl-dtl summary is also in accordance with the reference summary and sentence placement is also in accordance with the reference summary. Our best model also generate the summary without redundant information, since Devianti-Khodra-best-model has redundant information in second and third sentence. This proves that the dep-parser-srl-dtl model can improve the performance of the Devianti-Khodra-best-model model.

## V. CONCLUSION

Proposed multi-document summarization model for Indonesian news articles modify Devianti and Khodra architecture [1] by replacing SVOA extracting with SRL for Indonesian and replacing genetic algorithm with decision tree classification model to extract sentence features. The proposed model improves the performance of the existing best model for Indonesian news articles with genetic algorithm [1], by obtaining 0.313 average ROUGE-2 recall for 100 words summary and 0.394 average ROUGE-2 recall for 200 words summary. However, it still can not beat Devianti and Khodra best model that use default value for four sentence features weight [1].

Improvements can be made by adding training data to SRL model so it can recognize more predicate. Besides that, adding complex sentences into the SRL model also can be done since the SRL model can not recognize the right predicate in a complex sentence. Improvements also can be made by adding training data to the decision tree learning model.

ACKNOWLEDGMENT

We would like to thank Devianti and Khodra for providing us with news articles corpus. We also would like to thank Devina and Khodra for providing SRL corpus.

REFERENCES

[1] R. S. Devianti and M. L. Khodra, "Abstractive Summarization using Genetic Semantic Graph for Indonesian News Articles," International Conference on Advanced Informatics, Concepts, Theory, and Applications (ICAICTA), pp.

[2] A. Khan, N. Salim, and Y. G. Kumar, "A framework for multidocument abstractive summarization based on semantic role labeling," in Applied Soft Computing, vol. 30, February 2015.

[3] R. Reztaputra and M. L. Khodra, "Sentence structure-based summarization for Indonesian news articles," International Conference on Advanced Informatics, Concepts, Theory, and Applications (ICAICTA), pp. 1–6, August 2017.

[4] G. Garmastewira and M. L. Khodra, "Summarizing Indonesian News Articles Using Graph Convolutional Network," Journal of Information and Communication Technology, 18(3), pp. 345–365, July 2019.

[5] I. E. Devina and M. L. Khodra, "Semantic Role Labeling for Generating Template of Indonesian News Sentences," International Conference on Advanced Informatics, Concepts, Theory, and Applications (ICAICTA), pp.

[6] Hannah M.E., Geetha T.V., Mukherjee S. (2011) Automatic Extractive Text Summarization Based on Fuzzy Logic: A Sentence Oriented Approach. In: Panigrahi B.K., Suganthan P.N., Das S., Satapathy S.C. (eds) Swarm, Evolutionary, and Memetic Computing. SEMCCO 2011. Lecture Notes in Computer Science, vol 7076. Springer, Berlin, Heidelberg.

[7] Kupiec, J., Pedersen, J., & Chen, F. (1995). A trainable document summarizer. Proceedings of the 18th annual international ACM SIGIR conference on Research and development in information retrieval (pp. 68-73).